\documentclass[sigconf,nonacm]{acmart}

\usepackage{booktabs}
\usepackage{multirow}
\usepackage{amsmath}
\usepackage{enumitem}
\usepackage{array}
\usepackage{xcolor}
\usepackage{microtype}
\usepackage{graphicx}

\definecolor{bestcol}{HTML}{1a6e1a}

\title{Pseudo-Label Augmentation for Affect Sensing in Small
Collaborative Groups}

\author{Meisam Jamshidi Seikavandi}
\orcid{0000-0002-1271-2481}
\affiliation{%
  \institution{IT University of Copenhagen}
  \department{brAIn Lab}
  \city{Copenhagen}
  \country{Denmark}}
\affiliation{%
  \institution{GN Advanced Science, GN Hearing}
  \city{Ballerup}
  \country{Denmark}}
\email{meis@itu.dk}

\author{Tanya Ignatenko}
\affiliation{%
  \institution{GN Advanced Science, GN Hearing}
  \city{Ballerup}
  \country{Denmark}}

\author{Fabricio Batista Narcizo}
\orcid{0000-0003-1319-5153}
\affiliation{%
  \institution{IT University of Copenhagen}
  \city{K{\o}benhavn S}
  \country{Denmark}}
\affiliation{%
  \institution{GN Advanced Science, GN Hearing}
  \city{Ballerup}
  \country{Denmark}}
\email{narcizo@itu.dk}

\author{Paolo Burelli}
\orcid{0000-0003-2804-9028}
\affiliation{%
  \institution{IT University of Copenhagen}
  \department{brAIn Lab}
  \city{Copenhagen}
  \country{Denmark}}

\author{Jesper B{\"u}nsow Boldt}
\affiliation{%
  \institution{GN Advanced Science, GN Hearing}
  \city{Ballerup}
  \country{Denmark}}

\author{Andrew Burke Dittberner}
\orcid{0000-0003-4985-7083}
\affiliation{%
  \institution{GN Advanced Science, GN Hearing}
  \city{Glenview}
  \state{Illinois}
  \country{United States}}

\begin{document}

\begin{abstract}
Physiological affect sensing in naturalistic group interaction is typically limited more by label density than
by sensor density: wearable devices produce thousands of
time windows, while self-report labels are collected only a few times per
session.
This paper asks how to mine more supervisory signal from sparsely labelled
group sessions.
More broadly, we treat this as a step toward collective-state sensing:
better individual affect estimates within a team help reveal the affective
structure emerging at group level.
Using GroupAffect-4, a four-person collaborative dataset with wearable
physiology, eye tracking, Big Five personality, and post-task VAD labels, we
study a progression of label-mining strategies: no augmentation, Gaussian
Process (GP) pseudo-labelling, personality-aware trust weighting, and joint
personality-plus-confidence weighting under a shared target-construction
pipeline.
We further ask whether individual predispositions (here, Big Five
personality) together with group-level context improve which pseudo-labels
should be trusted.
They help, but not in the originally hypothesised graded way: the compressed
BFI cosine-similarity range ($0.91$--$0.99$) renders weight ordering
negligible (perm $p=0.39$), so personality similarity acts as a same-team
membership filter rather than a calibrated trust dial.
In the known-team setting, pseudo-label augmentation improves over the
labelled-only baseline, with the clearest benefit coming from within-team
filtering rather than from fine-grained personality ranking.
With smoothing, A1--A3 are tied to rounding on Valence and Arousal, while A3
has the highest Dominance score among the augmented SVM variants; the practical
choice is therefore dimension-dependent.
Cross-subject LOSO transfer remains encouraging for such a small and sparsely
labelled dataset, especially on Arousal, whereas strict session-isolated LOGO
removes the augmentation benefit.
At the current scale of only 10 groups, LOGO should be read as a conservative
lower bound on unseen-group transfer rather than as a definitive ceiling, but
the present evidence shows that personality weighting mainly operates as a
within-team mechanism.
\end{abstract}



\maketitle

\begin{center}
\small\textit{Author-prepared preprint. Accepted for publication in the ICMI 2026 Companion proceedings.}
\end{center}

\section{Introduction}
\label{sec:intro}

Wearable physiological sensors provide dense streams of electrodermal,
cardiac, oculomotor, pupil, and movement signals, but affect labels remain
sparse in naturalistic interaction.
In collaborative group studies, participants can only be interrupted for
self-report a few times per session.
This results in a structural mismatch, since collected datasets contain thousands of physiological windows with 
only a small number of labelled Valence--Arousal--Dominance (VAD) observations.
Typical solutions to this problem include  interpolation of labels over time or training larger
models on small labelled sets.
Both strategies are fragile.
Physiological affect is strongly person-specific, and the same arousal pattern
can reflect different appraisals depending on personality, task role, and group
context.
Therefore, we study the problem of extracting reliable pseudo-labels from sparsely labelled collaborative-group data that combines individual predispositions together with group-level structure.
To conduct the experiments, we use GroupAffect-4, which provides rich
multimodal group interaction signals but only sparse post-task affect reports.

In this paper, we investigate different label-mining strategies with sparse supervision, starting with a labelled-only baseline, following with GP-based temporal pseudo-labelling and
personality-aware trust weighting, and finally going into a joint
personality-plus-confidence variant.

One promising mechanism is personality-weighted pseudo-labelling: if some
participants are more similar to their groupmates in stable dispositions, then
their labels may transfer more reliably to neighbouring unlabelled windows or
to related within-team examples.
We therefore ask not only whether label augmentation helps, but which
augmentation assumptions hold under stricter transfer protocols.
The paper is intentionally scoped around low-capacity supervised models.
We use the same RBF-SVM throughout the main comparison so that improvements can
be attributed to pseudo-label construction rather than to architecture changes.
This is important because the dataset has only 103 labelled training windows in
the canonical split; high-capacity models can easily turn an augmentation
comparison into a capacity comparison.

We entered with a specific hypothesis: BFI cosine similarity might serve as a
stable trust signal for pseudo-label transfer, so that participants whose
trait profiles are more aligned with their group would yield more transferable
affective labels.
The data only partly support this idea.
Personality does help define which pseudo-labels are useful, but not through a
strong graded ranking of individuals.
Instead, within this dataset it behaves more like a same-team
\emph{membership filter} than like a finely calibrated trust dial.
That distinction matters for the paper's contribution: the value of
personality here lies in constraining the pseudo-label pool, not in providing
a precise interpersonal ordering.

The practical motivation is a known-team monitoring setting: a team completes
multiple collaborative tasks, physiology is recorded continuously, and only
sparse self-report is available, so earlier-task data from that same team can
support later-task prediction.
This is not the same as predicting a wholly unseen group, and much of the
paper is devoted to making that boundary explicit.
The canonical evaluation is intentionally demanding: the model is trained on
earlier tasks and evaluated on later, behaviorally distinct tasks, so success
reflects cross-task transfer within a known team rather than easier
within-task interpolation.
Viewed through the lens of collective-state modeling, this is a bottom-up
problem: group-level affective structure can only be tracked if
participant-level states are estimated robustly under sparse supervision.
Our contributions are:

\begin{itemize}[leftmargin=1.2em,itemsep=1pt,topsep=2pt]
  \item We define a leakage-aware augmentation taxonomy for sparse VAD labels
  in small groups, separating known-team temporal holdout from LOSO
  cross-subject and strict LOGO unseen-group stress tests.
  \item We compare three label-mining strategies under the same RBF-SVM base
  model: no augmentation, GP-based pseudo-labelling, and
  personality-mediated pseudo-label weighting, including a joint
  personality-plus-confidence variant.
  \item We show that personality-aware augmentation helps mainly as a
  within-team filtering mechanism: it improves the known-team setting, but its
  benefit comes from selecting useful pseudo-labels from the same team rather
  than from finely ranking individuals by trait similarity.
  \item We show that deployment regime matters: known-team augmentation is the
  main success case, LOSO transfer is encouraging at this dataset scale, and
  LOGO should currently be interpreted as a conservative lower bound on
  unseen-group transfer under limited group diversity.
\end{itemize}

\begin{sloppypar}
\section{Related Work}
\label{sec:related}

\paragraph{Dimensional affect and target construction.}
Russell's circumplex model foregrounds Valence and Arousal
\cite{russell1980circumplex}, while the PAD tradition adds Dominance as a
third interpersonal dimension \cite{mehrabian1996pleasure}.
The Self-Assessment Manikin (SAM) operationalises these axes as compact
visual self-reports \cite{bradley1994sam}, but ratings are ordinal, tied,
and often task-dependent.
Fixed threshold binning produces class imbalance that biases classifiers
\cite{he2009imbalanced,johnson2019classimbalance}; label distribution and
ordinal learning approaches argue for preserving ordered uncertainty
\cite{geng2016label,wen2023ordinal}, motivating our balanced
midpoint-tertile target pipeline.

\paragraph{Physiological and oculomotor affect recognition.}
DEAP~\cite{koelstra2012deap}, DREAMER~\cite{katsigiannis2018dreamer},
MAHNOB-HCI~\cite{soleymani2012mahnob}, WESAD~\cite{schmidt2018wesad},
and ASCERTAIN~\cite{subramanian2018ascertain} established multimodal
benchmarks; reviews confirm EDA, cardiac, and wearable signals are
useful but sensitive to preprocessing and splitting
protocol~\cite{shu2018review,ahmad2022survey,kreibig2010autonomic}.
MuMTAffect jointly models emotion and personality from multimodal
physiological signals~\cite{seikavandi2025mumtaffect}, while AFFEC combines
EEG, eye tracking, GSR, facial video, and personality and distinguishes felt
from perceived affect~\cite{seikavandi2026advancing}.
PPG and EDA processing follow NeuroKit2~\cite{makowski2021neurokit2} and
cvxEDA~\cite{greco2016cvxeda} respectively.

\paragraph{Social and group affect datasets.}
IEMOCAP~\cite{busso2008iemocap} and RECOLA~\cite{ringeval2013introducing}
introduced acted and remote dyadic affective interaction;
K-EmoCon~\cite{park2020kemocon} added naturalistic dyadic debate with
wearable sensors and multiple annotation perspectives.
AMIGOS~\cite{mirandacorrea2021amigos} explicitly studies affect, personality,
mood, and social context, but the group condition is shared media viewing
rather than collaborative problem solving.
Meeting and social-signal corpora such as AMI~\cite{mccowan2005ami},
SALSA~\cite{alameda2015salsa}, MatchNMingle~\cite{cabrera2018matchnmingle},
and ELEA~\cite{sanchez2012emergent} capture rich group behaviour and
social signals.
GroupAffect-4~\cite{groupaffect4} occupies the remaining intersection:
structured four-person collaboration with per-person physiology, eye
tracking, VAD self-report, and pre-session personality.

\paragraph{Sparse labels, augmentation, and generalisation.}
Small labelled physiological datasets make model selection unusually
dependent on target quality and validation design.
Time-series augmentation can improve wearable classifiers
\cite{um2017data,iwana2021augmentation}, and contrastive methods such as
TS-TCC~\cite{eldele2021tscc} learn representations from unlabelled
segments.
Semi-supervised consistency methods, from Mean
Teacher~\cite{tarvainen2017mean} to FixMatch~\cite{sohn2020fixmatch},
show that pseudo-labels are useful only when confidence is reliable;
pseudo-labelling adds unlabelled examples above a confidence threshold
\cite{lee2013pseudo}.
Gaussian Processes provide uncertainty-aware interpolation in label
space~\cite{rasmussen2006gaussian,atcheson2017gp}, complementing
confidence-threshold approaches.
Subject- and group-aware splits are essential for interpreting wearable
affect model generalisation~\cite{ahmad2022survey,ventura2022validation}.

\paragraph{Personality and affect dynamics.}
The Big Five~\cite{john1999bigfive} is the dominant personality taxonomy;
Neuroticism and Extraversion are primary affect predictors
\cite{watson1992traits}.
ASCERTAIN and AMIGOS show that personality shapes physiological affect
responses~\cite{subramanian2018ascertain,mirandacorrea2021amigos}.
Personality-aware face-to-face modelling has also found that combining
temporal gaze, stimulus context, and Big Five traits particularly benefits
felt-emotion prediction~\cite{seikavandi2025modelling}.
\citet{meng2026moderating} used CTSEM to model how all five traits moderate
VAD dynamics during real interaction, finding trait-specific moderation
timescales that directly motivate our per-dimension personality trust
weighting while also warning that trait effects are contextual and slow
relative to 15-second windows.
Emotional contagion suggests group members converge in affective state
\cite{hatfield1993emotional}, supporting cross-person pseudo-label transfer
weighted by personality similarity.

\paragraph{Why personality and not only physiological synchrony?}
Physiological synchrony is measured at the same timescale as the target
windows, but it is ambiguous: co-fluctuation can reflect a shared external
event, concurrent movement, or genuine affective convergence.
Personality similarity is slower and more stable, providing a durable prior
for weighting cross-person pseudo-labels when self-reports are sparse and
uncertain.
Our contribution is to operationalise BFI cosine similarity as a trust
weight and to test its deployment boundary under known-team, LOSO, and
strict LOGO protocols.
\end{sloppypar}

\section{Dataset and Targets}
\label{sec:dataset}

GroupAffect-4 contains 10 four-person collaborative sessions, with 40
participants retained after data completeness checks
\cite{groupaffect4,seikavandi2026affectai}.
Each session contains five sequential tasks: T0 passive rest, T1 information
pooling, T2 negotiation, T3 ideation, and T4 public-goods interaction.
After each task, participants reported Valence, Arousal, and Dominance using
9-point SAM scales \cite{bradley1994sam}; T4 Dominance was not administered
by design.
Participants also completed the BFI-44 Big Five Inventory
\cite{john1999bigfive} before the session.

\paragraph{Physiology and eye-tracking inputs.}
Table~\ref{tab:modalities} summarises the two wearable hardware platforms.
All streams are time-stamped via LSL (Lab Streaming Layer) for cross-modal
alignment and yield a 49-dimensional summary feature vector per 15-second
window (mean, standard deviation, slope, and percentile statistics per
channel).
PPG is processed with NeuroKit2~\cite{makowski2021neurokit2}; EDA is
decomposed into phasic (SCR) and tonic (SCL) components via
cvxEDA~\cite{greco2016cvxeda}; gaze/pupil statistics follow standard
fixation and pupillometry conventions~\cite{laeng2012pupillometry}.

\begin{table}[t]
\caption{Physiology and eye-tracking inputs (GroupAffect-4).}
\label{tab:modalities}
\centering
\setlength{\tabcolsep}{3pt}
\footnotesize
\begin{tabular}{>{}l >{}l >{}l >{}l}
\toprule
\textbf{Modality} & \textbf{Device} & \textbf{Rate} & \textbf{Features} \\
\midrule
PPG        & EmotiBit & $\sim$25\,Hz & HR, RMSSD, SDNN, amp. \\
EDA        & EmotiBit & $\sim$25\,Hz & SCL/SCR, phasic, tonic \\
Skin temp. & EmotiBit & $\sim$25\,Hz & Mean, slope \\
IMU        & EmotiBit & $\sim$25\,Hz & Accel.~mag., movement \\
Gaze       & Tobii G3 & 40--60\,Hz   & Fix.~rate, saccade \\
Pupil      & Tobii G3 & 40--60\,Hz   & Mean diam., dilation var. \\
\bottomrule
\end{tabular}
\end{table}

We use 15-second windows with 50\% overlap; the ablation
(\S\ref{sec:preprocessing}) shows that 30-second pooling improves Arousal and
partly Dominance but weakens Valence and reduces the labelled set. This scale
is also consistent with prior affect work~\cite{park2020kemocon,
prabhu2024collective,kreibig2010autonomic}. Each post-task SAM rating is
propagated to its task windows, assuming approximate within-task stationarity;
these are task-level supervisory proxies, not dense evidence of fine-grained
trajectories. Overlapping windows are not statistically independent, so we
make no inferential claims that treat them as independent observations.

The canonical split is a known-team temporal holdout:
train on T0+T1 ($N=103$ windows), validate on T2 ($N=78$), and test on T3
($N=65$).
This forward, cross-task ordering mirrors deployment in which earlier-task
observations support later-task prediction within the same team. T3 provides
complete VAD labels for 32 participants; 8 lack usable T3 physiology. Because
T4 Dominance was not administered, T4 appears only in auxiliary task-CV with
that dimension omitted.
Sliding 15-second windows with 50\% overlap yield 292 labelled windows
(from T0--T4) and 8{,}221 unlabelled windows drawn from between-task
intervals and portions of tasks without SAM labels.
We also report LOSO and LOGO stress tests for cross-person and cross-group
generalisation.
All results report macro-F1 (sklearn average=\texttt{`macro'}) separately for
V/A/D because their distributions and responses to augmentation differ.

\paragraph{Target construction.}
Raw SAM scores are converted into Low, Mid, and High classes using midpoint
tertile thresholds fitted on training data only.
Boundaries are set at midpoints between integer response values.
This keeps tied Likert clusters in the same class.
Per-dimension inverse-frequency weights correct residual imbalance.
This target pipeline replaces a fixed-threshold baseline that inflated
Valence F1 when the T3 test set contained no Low-valence examples.
The same fitted thresholds are used to discretise GP posterior moments for
augmented windows, so hard labels and pseudo-labels live in the same class
space.

\subsection{Why Target Construction Matters}
\label{sec:target_construction}

The label pipeline is part of the contribution because augmentation quality
cannot be evaluated independently of target quality.
The original 1--9 SAM responses are ordinal, tied, and task-dependent.
For example, T3 ideation contains many positive-valence reports; fixed
thresholds such as Low $\leq 3$, Mid $=4$--$6$, and High $\geq 7$ can remove
one class from the test fold.
When that happens, macro-F1 is no longer a three-class metric even if the code
reports a single number.

Balanced midpoint-tertile thresholds avoid this failure mode.
The split points are fitted on training labels only, then moved to midpoints
between integer response values so that all tied Likert responses stay in the
same class.
This gives a small but honest benchmark: the SVM baseline becomes stronger
because class semantics match the task distribution, while the MLP no longer
benefits from silent class collapse.
All augmentation variants in this paper are evaluated after this correction.

\begin{table}[t]
\caption{Label-pipeline ablation motivating the shared target construction.
Values are T3 macro-F1 for an MLP without augmentation.}
\label{tab:target_ablation}
\centering
\setlength{\tabcolsep}{4pt}
\small
\begin{tabular}{lccc}
\toprule
\textbf{Label pipeline} & \textbf{V} & \textbf{A} & \textbf{D} \\
\midrule
Fixed thresholds, no weights & 0.575 & 0.516 & 0.338 \\
Balanced tertile, no weights & 0.372 & 0.443 & 0.429 \\
Full target pipeline         & 0.405 & 0.464 & 0.495 \\
\bottomrule
\end{tabular}
\end{table}

\begin{table}[t]
\caption{Canonical benchmark setup used throughout this paper.}
\label{tab:setup}
\centering
\setlength{\tabcolsep}{4pt}
\small
\begin{tabular}{ll}
\toprule
\textbf{Component} & \textbf{Definition} \\
\midrule
Dataset & GroupAffect-4, 40 participants, 10 groups \\
Signals & Gaze, pupil, EDA, PPG, skin temp., IMU \\
Features & 49 physiology/eye-tracking summary features \\
Labels & Balanced midpoint-tertile VAD classes \\
Train/val/test & T0+T1 / T2 / T3 \\
Base model & RBF-SVM, $C=1$, balanced class weights \\
Metric & Macro-F1 per VAD dim.\ (V/A/D reported separately) \\
\bottomrule
\end{tabular}
\end{table}

\subsection{Deployment Regimes}
\label{sec:deployment}

We distinguish three regimes because ``generalisation'' has different meanings
in group affect sensing.
\textbf{Known-team temporal holdout} asks whether earlier observations from a
team can help predict that same team's later task windows.
This is realistic for longitudinal team-monitoring systems and is the primary
setting for A2/A3.
In the canonical split, this means transferring from T0+T1 to later tasks
rather than re-predicting the same task type, which makes the setting
substantively harder because negotiation and ideation induce different group
dynamics and physiological context.
\textbf{LOSO} asks whether models transfer to an unseen participant within the
same task family.
\textbf{LOGO} asks whether models transfer to an entirely unseen group, where
all members, baselines, and interaction dynamics are absent from training.
The last regime is much harder and should not be conflated with the first.

This separation is important for personality-based augmentation.
BFI-44 is collected before the session, so the personality similarity weight is
not label leakage.
However, a personality weight applied to a pseudo-label does not make the
pseudo-label itself automatically leakage-free.
The strict LOGO experiment therefore regenerates the pool while excluding the
held-out group entirely.

\section{Augmentation Methods}
\label{sec:methods}

All augmentation variants use the same SVM base model.
Unlabelled windows are pseudo-labelled and assigned sample weights; test
evaluation always uses ground-truth SAM-derived labels only.
Let $\mathcal{D}_{L}=\{(\mathbf{x}_i,y_i,w_i=1)\}$ denote labelled training
windows and $\mathcal{D}_{U}=\{(\mathbf{x}_u,\hat{y}_u,\alpha_u)\}$ denote
accepted pseudo-labelled windows.
The SVM is trained on
\begin{equation}
  \mathcal{D}_{aug} = \mathcal{D}_{L} \cup \mathcal{D}_{U},
\end{equation}
with sample weight $\alpha_u \in [0,1]$ for pseudo-labelled examples.
For A2, $\alpha_u$ is the BFI similarity weight; for A3 it is the product of
BFI similarity and GP posterior confidence.
No test labels are used in pseudo-label generation.

\paragraph{GP pseudo-labels.}
For each participant and VAD dimension, we fit an Ornstein--Uhlenbeck
Gaussian Process (GP) to
task-level SAM observations.
In the canonical split this provides only T0 and T1 anchors per participant;
the GP is therefore used as a deliberately weak interpolation prior, not as
evidence that a continuous affect trajectory has been recovered.
The posterior over the 1--9 SAM scale is integrated across the
training-fitted midpoint thresholds to produce a three-class soft label.
For the SVM, the pseudo-label class is the posterior argmax and the
posterior confidence becomes a sample weight.
Threshold alignment is critical: the GP posterior is first expressed on the
original 1--9 SAM scale and only then integrated over the training-fitted
Low/Mid/High thresholds.
Using stale thresholds from a different target pipeline can invert the effect
of augmentation because the pseudo-label class prior no longer matches the
supervised labels.

\paragraph{Personality trust weights.}
When a pseudo-label is generated from participant $j$'s physiology, we weight
it by how similar $j$ is to the rest of the group:
\begin{equation}
  \mathrm{BFI\_sim}_j =
  \frac{1}{|G|-1}\sum_{k \neq j}
  \frac{\mathbf{b}_j \cdot \mathbf{b}_k}{\|\mathbf{b}_j\|\|\mathbf{b}_k\|}.
\end{equation}
Here $G$ is the four-person group, $\mathbf{b}_j$ is participant $j$'s
BFI-44 trait vector, and the sum runs over $j$'s groupmates $k$.
Importantly, this is a \emph{participant-level} weight: it measures how much
the pseudo-label \emph{source} $j$ resembles the group as a whole, not a
pairwise weight between source $j$ and a specific recipient.
Every pseudo-label generated from $j$'s unlabelled windows receives the same
$\mathrm{BFI\_sim}_j$ regardless of which recipient's model is being trained.
A2 uses $\mathrm{BFI\_sim}_j$ as the sample weight $\alpha_u$ for every
pseudo-label produced from participant $j$'s unlabelled windows:
a participant who is more similar to the group on average is assumed to share
affective appraisals with groupmates, making their pseudo-labels more
trustworthy as training examples for any group member's model.
A3 multiplies this personality weight by GP posterior confidence.
The BFI weight is leakage-free because personality is collected before
the session; the pseudo-label class itself inherits only information from
training-task SAM observations (Scenario~1) or fully held-out-group-excluded
GP posteriors (Scenario~2).

\paragraph{Leakage taxonomy.}
Scenario~1 is the known-team temporal holdout: the GP is fitted on T0+T1
task-level SAM observations only; T2, T3, and T4 labels are withheld from GP
fitting, so the pseudo-label parameters contain no information from the
validation or test tasks.
Pool windows are drawn from between-task intervals in T0 and T1.
Scenario~2 is session-isolated LOGO: the held-out session is excluded from
GP fitting, pool windows, BFI maps, and parameter estimation entirely.
Only Scenario~2 tests unseen-group deployment.

\paragraph{Pool filtering.}
Pool windows are filtered by task and source self-report index.
For the known-team temporal split, T3 labels are excluded from augmentation
because T3 is the test task.
For LOGO, all windows from the held-out group are removed before GP fitting,
BFI-map construction, and pseudo-label acceptance.
This makes LOGO a stricter test than simply training on non-test labelled
windows while leaving the held-out group's unlabelled pool in the augmentation
set.

\begin{table}[t]
\caption{Checks required before claiming leakage-aware augmentation.}
\label{tab:checks}
\centering
\setlength{\tabcolsep}{3pt}
\small
\begin{tabular}{>{\raggedright\arraybackslash}p{0.34\columnwidth}>{\raggedright\arraybackslash}p{0.54\columnwidth}}
\toprule
\textbf{Check} & \textbf{Why it matters} \\
\midrule
Train-fitted thresholds & Keeps hard and pseudo-labels in one class space. \\
Exclude test-task labels & Prevents direct temporal label leakage. \\
Regenerate LOGO pool & Removes held-out group from all augmentation signals. \\
Keep BFI out of SVM features & Preserves personality as trust, not identity. \\
Report V/A/D separately & Reveals that A2 mainly helps V and D. \\
\bottomrule
\end{tabular}
\end{table}

\begin{table}[t]
\caption{Main augmentation variants carried through the paper.
Preprocessing and window-size side ablations are reported separately in
Section~\ref{sec:preprocessing}.}
\label{tab:variant_defs}
\centering
\setlength{\tabcolsep}{3pt}
\small
\begin{tabular}{l>{\raggedright\arraybackslash}p{0.57\columnwidth}}
\toprule
\textbf{Variant} & \textbf{Pseudo-label / weight rule} \\
\midrule
A0 & No augmentation; labelled T0+T1 windows only. \\
A1 & GP class argmax for all VAD dimensions. \\
A2 & GP class argmax, BFI cosine similarity as trust weight. \\
A3 & GP class argmax, BFI similarity multiplied by GP confidence. \\
\bottomrule
\end{tabular}
\end{table}

\subsection{Why We Tested Personality Weighting}
\label{sec:why_bfi}

The A2 design deliberately avoids using personality as a high-capacity neural
conditioning input.
At $N=103$, adding trait-conditioned projection matrices gives the model more
ways to overfit than the data can support.
Instead, personality plays a weaker role: it adjusts how much a pseudo-label
should be trusted.

\textbf{Prior hypothesis.}~Agreeableness, Extraversion, and Neuroticism shape
how participants appraise cooperation, conflict, and social pressure.
We therefore hypothesised that a pseudo-label source whose traits are more
centrally located relative to the group (higher $\mathrm{BFI\_sim}_j$)
would generate more transferable labels, and that this group-centrality weight
would drive the A2 gain.
A2 tests this hypothesis by using $\mathrm{BFI\_sim}_j$ as the sole
pseudo-label weight.

\paragraph{Empirical test and finding.}
Section~\ref{sec:uncertainty} reports a permutation test in which BFI cosine
similarities are randomly shuffled $500$ times and A2 is re-run with each
shuffled weight vector.
The result is that shuffled weights yield a per-dim F1 pattern essentially
indistinguishable from real weights on every dimension (perm $p = 0.39$).
The weight \emph{ordering} therefore explains little of the gain.
An analysis of the BFI cosine-similarity range in this dataset shows why:
all group members have compressed similarity scores ($0.91$--$0.99$), so
the weight variation across pseudo-labels is negligible.
In this dataset, BFI functions as a group-membership filter (accepting pool
windows from the known teams recorded in the dataset) rather than a precision
group-centrality weight that ranks individuals by proximity to the group
centre.
This reframes the A2 contribution: the within-team per-dim gains on Valence
and Dominance are real, but they come from having more pseudo-labels, not
from personality-specific ordering.
Because we did not include an A1 control with a constant pseudo-label weight
near the observed BFI range (e.g., $\alpha=0.95$), the current results cannot
separate the near-uniform scalar weighting effect from the effect of admitting
the same-team pseudo-label pool. This control is required before attributing
any residual gain specifically to personality.
Validating the graded group-centrality hypothesis requires groups with a wider
BFI distribution or deliberate inclusion of personality-outlier participants.

\paragraph{Capacity control.}
The SVM does not receive BFI features directly.
This prevents the model from memorising participant or group-level trait
patterns.
Personality only enters through sample weights on pseudo-labelled windows.
Even when those weights are near-uniform (compressed BFI range), this
constraint keeps the augmentation interpretable.

\section{Results}
\label{sec:results}

\subsection{Known-Team Augmentation}
\label{sec:known_team}

\begin{figure}[t]
\centering
\includegraphics[width=\columnwidth]{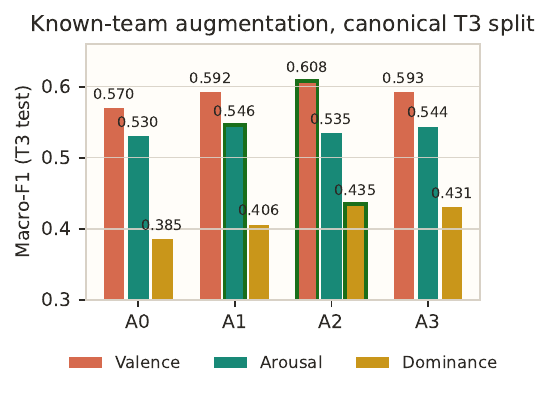}
\caption{SVM augmentation variants on the canonical T3 test split (A0: no
augmentation; A1: GP-only; A2: BFI-only; A3: BFI$\times$GP). Green outlines
mark the best variant per VAD dimension.}
\Description{Grouped bar chart of macro-F1 for A0--A3 on Valence, Arousal,
and Dominance. A2 is highest for Valence and Dominance, while A1 is highest
for Arousal.}
\label{fig:svm_aug}
\end{figure}

Figure~\ref{fig:svm_aug} reports the main result of the paper.
In the known-team setting, mining additional labels from unlabelled windows is
useful, and the clearest benefit comes from the personality-aware within-team
filter rather than from GP interpolation alone.
A2 gives the strongest gains on Valence (0.608) and Dominance (0.435), while
Arousal peaks under A1 (0.546) and is more responsive to temporal context than
to personality-aware weighting.

The pattern is easiest to understand by VAD dimension.
Valence and Dominance are more appraisal-heavy and relational, so they are the
dimensions where within-team label transfer helps most.
Arousal is more phasic and event-driven, so the same personality cue provides
less leverage before additional temporal smoothing is introduced.
At the same time, the benefit should not be over-interpreted as evidence of
fine-grained personality ranking.
As shown later by the permutation analysis
(\S\ref{sec:uncertainty}), the gain is driven more by same-team pseudo-label
availability than by the exact ordering of BFI weights.
At this dataset scale, the known-team improvements are therefore best treated
as informative point estimates rather than as statistically settled effect
sizes.

\subsection{LOSO Cross-Subject Stress Test}
\label{sec:loso}

To separate cross-person from cross-group transfer, we additionally report a
LOSO stress test under the same balanced target-construction pipeline.
Each fold holds out one participant's T3 windows and aggregates results across
the 32 participants with T3 labels.
Following strict cross-subject isolation, no pseudo-label augmentation is used
in this stress test.
LOSO remains clearly harder than the known-team setting, especially on
Valence, but it is not devoid of transferable signal.
Against the uniform three-class reference of macro-F1 $\approx 1/3$, Arousal
is above that reference, Valence is below it, and Dominance (0.334) is
effectively at it.
For a dataset of only 40 retained participants and sparse labels, this makes
the LOSO result better read as encouraging than disappointing: the model is
not solving cross-subject transfer, but it is already extracting some
participant-independent affective structure under strict participant
isolation.
LOSO therefore acts as an intermediate regime between the strong known-team
setting and the harder unseen-group LOGO setting.

\begin{table}[t]
\caption{LOSO cross-subject stress test (no augmentation).
Aggregate over 32 held-out participants with T3 labels.}
\label{tab:loso}
\centering
\setlength{\tabcolsep}{4pt}
\small
\begin{tabular}{lccc}
\toprule
\textbf{Variant} & \textbf{V} & \textbf{A} & \textbf{D} \\
\midrule
A0: LOSO no aug & 0.190 & 0.438 & 0.334 \\
\bottomrule
\end{tabular}
\end{table}

\subsection{Strict Session Isolation}
\label{sec:logo}

\begin{figure}[t]
\centering
\includegraphics[width=\columnwidth]{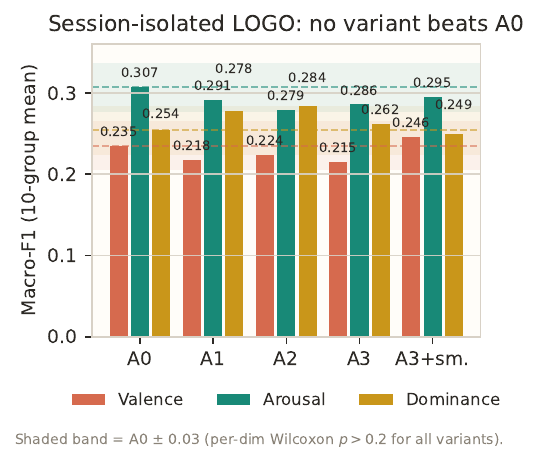}
\caption{Session-isolated LOGO augmentation (\textbf{leakage-clean}: pool, GP,
and BFI maps all exclude the held-out session), mean across 10 held-out
groups. The shaded band marks A0 $\pm$ 0.03 per dimension; every augmented
variant falls inside it, and no variant significantly outperforms A0
(Wilcoxon $p{>}0.2$ for all). Per-dim only; no pooling.}
\Description{Grouped bar chart of session-isolated LOGO macro-F1 across ten
held-out groups. All augmented variants remain within 0.03 of the A0 baseline
for each VAD dimension.}
\label{fig:logo}
\end{figure}

Figure~\ref{fig:logo} shows that the A2 gain is scoped.
When the held-out group is excluded from all pool generation, augmentation no
longer improves performance.
This does not invalidate A2; it identifies its deployment regime.
Personality weighting is useful for a known team with earlier-task data, not a
general cross-group transfer mechanism at the current 10-group scale.

At the same time, LOGO should be interpreted carefully.
This is the strictest and most data-hungry evaluation regime in the paper:
the model must bridge between-group differences in baseline physiology,
interaction style, and task interpretation with only a small number of
training groups.
Under those conditions, the present LOGO scores are best read as a
conservative lower bound on unseen-group transfer rather than as evidence that
cross-group label mining is intrinsically ineffective.
Larger and more diverse group datasets may narrow this gap, but that remains a
hypothesis for future work rather than a claim established here.

\subsection{Uncertainty and Significance}
\label{sec:uncertainty}

The canonical T0+T1$\rightarrow$T3 comparison is a fixed split with
deterministic SVM training, so Figure~\ref{fig:svm_aug} reports point
estimates rather than seed variance.
\paragraph{Multi-fold task-CV.}
We run A0, A2, and A3 across all five held-out tasks (T0--T4) as an exploratory
check of single-fold stability. A1 was not included in this auxiliary run, so
this analysis does not provide a complete four-variant comparison.
The multi-fold view supports the same broad story as the canonical split:
augmentation is most helpful on Valence and Dominance, while Arousal is less
consistently affected at this preprocessing level.
Between-fold variance remains substantial because the held-out tasks differ in
interaction style and difficulty; T3 is comparatively favourable, whereas more
adversarial tasks are harder.
\paragraph{Permutation test on BFI weighting.}
To test whether the \emph{personality-specific ordering} of trust weights
drives the A2 gain (\S\ref{sec:why_bfi}), we shuffle the BFI
cosine-similarity scores across pool windows 500 times while keeping
pseudo-labels unchanged.
The shuffled-weight null distribution is essentially indistinguishable from
the real-weight result (perm $p = 0.39$), confirming that weight ordering is
not what drives the gain.
\paragraph{LOGO cross-group test.}
For protocol-level uncertainty, LOGO is evaluated across 10 held-out
sessions (Figure~\ref{fig:logo}); per-dim deltas of every augmentation
variant against A0 are within $|0.030|$ on Valence, Arousal, and Dominance
separately.
Paired Wilcoxon signed-rank tests on per-session per-dim F1 show no
significant difference against A0 (all $p{>}0.2$).
At the current dataset scale, this means only that unseen-group benefit is not
yet established.
The safest interpretation is that the present A2 gain operates mainly within
the known-team regime, while LOGO remains a conservative empirical boundary
under limited group coverage.

\subsection{Preprocessing and Window-Size Ablation}
\label{sec:preprocessing}

As a side ablation rather than a second main story, we test whether two
upstream changes alter the main augmentation pattern:
(i) centered temporal smoothing of EDA and PPG only (45-second context
window, preserving gaze/pupil/IMU), and
(ii) 30-second pooled windows (consecutive 15-second pairs averaged;
approximate).
Table~\ref{tab:preproc} reports the resulting known-team T3 scores.

\begin{table}[t]
\caption{Side ablation on smoothing and window size (known-team, test = T3).
Per-dim only; no pooling.}
\label{tab:preproc}
\centering
\setlength{\tabcolsep}{3.5pt}
\small
\begin{tabular}{llccc}
\toprule
\textbf{Condition} & \textbf{Variant} & \textbf{V} & \textbf{A} & \textbf{D} \\
\midrule
\multirow{4}{*}{15s baseline}
 & A0             & 0.570 & 0.530 & 0.385 \\
 & A1             & 0.592 & 0.546 & 0.406 \\
 & A2 (best aug)  & 0.608 & 0.535 & 0.435 \\
 & A3             & 0.593 & 0.544 & 0.431 \\
\midrule
\multirow{4}{*}{EDA/PPG smooth (ctr.)}
 & A0             & 0.601 & 0.566 & 0.409 \\
 & A1             & 0.642 & \textbf{0.624} & 0.427 \\
 & A2             & \textbf{0.643} & \textbf{0.624} & 0.420 \\
 & A3             & 0.642 & \textbf{0.624} & \textbf{0.442} \\
\midrule
30s windows (A0)  & A0             & 0.477 & 0.662 & 0.452 \\
\bottomrule
\end{tabular}
\end{table}

This side analysis changes the practical interpretation more than the
mechanistic one.
First, better temporal preprocessing helps every retained variant, with the
strongest effect on Arousal.
Second, smoothing makes A3 competitive within the fixed 15-second SVM
comparison because better-behaved physiology improves the usefulness of GP
confidence; however, A1--A3 are tied to rounding on Valence and Arousal, and
A3's Dominance margin over A1 is 0.015 on 65 test windows.
Third, the 30-second ablation does not overturn the 15-second design choice.
Longer windows improve Arousal and partly Dominance under the A0 baseline, but
they also weaken Valence, halve temporal resolution, and reduce the effective
number of labelled training examples.
We therefore retain 15-second windows as the main setting because they are the
better all-around compromise for per-dimension comparison and for testing
augmentation under severe label scarcity.

\section{Mechanistic Interpretation}
\label{sec:mechanism}

The dimensions respond differently. Valence and Dominance depend more on
stable appraisal and relational context, explaining their larger A2 gains;
Arousal is more phasic and benefits more from smoothing or longer windows.
Dominance remains difficult because physiology omits speaking, directed gaze,
and influence structure. A3 can prune too aggressively without smoothing, but
becomes competitive once EDA/PPG smoothing improves GP confidence. The narrow
BFI range nevertheless supports only a same-team filter, not calibrated
personality ranking.

\section{Discussion}
\label{sec:discussion}

\paragraph{Personality as a filter, not a calibrated trust signal.}
Direct BFI conditioning overfits at the current label density, so A2 uses
personality only as a pseudo-label trust weight (\S\ref{sec:why_bfi}) --- a
lower-capacity role better matched to $N=103$.
As the permutation test and compressed similarity range
(\S\ref{sec:uncertainty}) show, this weight acts as a binary selector (keep
vs.\ discard pool windows) rather than a graded confidence dial; validating a
truly graded version requires groups with more diverse personality profiles.
\paragraph{Known-team benefit is still useful.}
Many practical group-sensing deployments observe the same team over multiple
tasks or sessions.
In that setting, A2 is legitimate: it uses pre-session personality and
earlier-task unlabelled windows to smooth the team's affect trajectory.
The important wording is ``same team.''
The A2 result should not be sold as generalisation to strangers; it is a
within-team calibration method for sparse labels.
That narrower claim is still useful because many collaborative systems care
about monitoring an ongoing team rather than classifying arbitrary new groups.
\paragraph{Practical takeaway.}
The results do not establish one universally best recipe. Within the fixed
15-second smoothed SVM comparison, A1--A3 tie to rounding on Valence and
Arousal, while A3 has the highest augmented Dominance score. The no-augmentation
MLP target-pipeline ablation (D$=0.495$) and the 30-second A0 ablation
(A$=0.662$, D$=0.452$) score higher on some dimensions, but change architecture
or windowing and are therefore not controlled augmentation comparisons.
Selection should follow the target dimension and deployment regime: A2 is the
more interpretable within-team filter, A3 is a conditional option with smoothed
physiology, and A0 remains the conservative unseen-group baseline.

\paragraph{Cross-subject and cross-group transfer.}
LOSO is encouraging for such a small and sparsely labelled dataset, especially
on Arousal, whereas LOGO remains below known-team performance. With only 10
groups, LOGO is a lower bound rather than evidence that cross-group transfer
is intrinsically ineffective. New groups may require brief self-report
calibration before augmentation, which remains untested.
\paragraph{Relational Dominance.}
Dominance is relational, yet the present tertile thresholds are fitted
globally and the four members' labels enter the model as separate samples.
The method therefore does not exploit within-group rank, reciprocity, or
influence structure. Future work should model these relations explicitly with
speaker turns, person-directed gaze, and group-relative Dominance targets.
\section{Limitations}
\label{sec:limitations}

The dataset contains only 10 groups, so strict LOGO estimates have high fold
variance.
The canonical A2 comparison should be interpreted as a temporal known-team
result, not as unseen-group deployment.
Future work should add clustered confidence intervals, evaluate repeated
sessions per group, and test calibration strategies for new teams.
Because LOGO is evaluated on a small number of groups, we cannot conclude from
this dataset alone how close unseen-group performance could move to known-team
results under broader group coverage.
The reported A2 improvement is small in absolute terms and should be treated
as a calibrated low-label gain rather than a large general-purpose improvement.
Because windows overlap within participants and groups, future statistical
tests should resample at participant or session level rather than assuming
independent windows.
The pseudo-labels are not validated against dense human annotation or an
independent affect measure, and retained pseudo-label counts per variant are
not reported here; both limit verification of the augmentation mechanism.
The canonical positive result also relies on one fixed T0+T1$\rightarrow$T3
split, with T3 comparatively favourable, while the auxiliary task-CV omits
A1. The cross-task findings should therefore be treated as exploratory until
all variants are compared across repeated task and group splits.
Several explored directions also did not become stable mainline methods at
this dataset scale.
Direct trait-conditioned modelling added too much capacity for $N=103$ and was
more prone to overfitting than the lighter A2 weighting scheme; in the
unsmoothed setting, A3's extra GP-confidence gate could also prune too
aggressively, while 30-second windows improved Arousal, and partly Dominance,
at the cost of Valence and temporal resolution.
These negative results are still informative: they suggest that, for now,
simple known-team augmentation is more defensible than stronger claims about
rich personalization or immediate unseen-group transfer.

\section{Conclusion}
\label{sec:conclusion}

This paper frames sparse affect annotation as a label-mining problem: in group
interaction datasets, the bottleneck is not sensor volume but the scarcity of
trustworthy supervisory signal.
Across a progression from no augmentation, to GP pseudo-labelling, to
personality-aware weighting, experiments on GroupAffect-4 show that
additional supervision can be mined usefully in the known-team regime, a
bottom-up contribution toward collective-state sensing in small teams.

Our original hypothesis --- that BFI cosine similarity would rank
pseudo-label trustworthiness by psychological proximity --- was falsified:
the compressed BFI range ($0.91$--$0.99$) renders weight ordering negligible
(perm $p=0.39$), so personality functions as a same-team membership filter
rather than a calibrated trust dial.
A2 still delivers point-estimate V/D gains ($+0.038$ / $+0.050$), because
pseudo-label quantity from recorded teams matters even when weight ordering
does not.

With centered EDA/PPG smoothing, A3 is competitive within the fixed 15-second
SVM comparison (V$=0.642$ / A$=0.624$ / D$=0.442$), but the appropriate
configuration remains dimension- and deployment-dependent.
LOSO is encouraging for such a small dataset, while LOGO, at only 10 groups,
should be read as a conservative lower bound on unseen-group transfer.
Label augmentation therefore helps sparse-label physiological affect
modelling, but its benefits depend strongly on deployment regime and on how
individual predispositions and group context constrain pseudo-label trust.

\bibliographystyle{ACM-Reference-Format}
\bibliography{icmi2026_refs}

@article{groupaffect4,
  title     = {{GroupAffect-4}: A Multimodal Dataset of Four-Person
               Collaborative Interaction},
  author    = {Seikavandi, Meisam Jamshidi and Modica, Alice and
               Obara, Anna and Shaffi, Shan Ahmed and
               Narcizo, Fabricio Batista and Ignatenko, Tanya and
               Vucurevich, Ted and Haddad, Karim and Barratt, Daniel and
               Overholt, Daniel and others},
  journal   = {arXiv preprint arXiv:2605.19765},
  year      = {2026}
}

@article{katsigiannis2018dreamer,
  author    = {Katsigiannis, Stamos and Ramzan, Naeem},
  title     = {{DREAMER}: A Database for Emotion Recognition Through {EEG}
               and {ECG} Signals from Wireless Low-cost Off-the-Shelf Devices},
  journal   = {IEEE Journal of Biomedical and Health Informatics},
  year      = {2018},
  volume    = {22},
  number    = {1},
  pages     = {98--107},
  doi       = {10.1109/JBHI.2017.2688239}
}

@article{koelstra2012deap,
  author    = {Koelstra, Sander and Muhl, Christian and Soleymani, Mohammad
               and Lee, Jong-Seok and Yazdani, Ashkan and Ebrahimi, Touradj
               and Pun, Thierry and Nijholt, Anton and Patras, Ioannis},
  title     = {{DEAP}: A Database for Emotion Analysis Using Physiological
               Signals},
  journal   = {IEEE Transactions on Affective Computing},
  year      = {2012},
  volume    = {3},
  number    = {1},
  pages     = {18--31},
  doi       = {10.1109/T-AFFC.2011.15}
}

@inproceedings{ringeval2013introducing,
  author    = {Ringeval, Fabien and Sonderegger, Andreas and Sauer, Juergen
               and Lalanne, Denis},
  title     = {Introducing the {RECOLA} Multimodal Corpus of Remote
               Collaborative and Affective Interactions},
  booktitle = {Proc.\ 10th IEEE Int.\ Conf.\ on Automatic Face and
               Gesture Recognition (FG)},
  year      = {2013},
  pages     = {1--8},
  doi       = {10.1109/FG.2013.6553805}
}

@article{busso2008iemocap,
  author    = {Busso, Carlos and Bulut, Murtaza and Lee, Chi-Chun and
               Kazemzadeh, Abe and Mower, Emily and Kim, Samuel and
               Chang, Jeannette~N. and Lee, Sungbok and
               Narayanan, Shrikanth~S.},
  title     = {{IEMOCAP}: Interactive Emotional Dyadic Motion Capture
               Database},
  journal   = {Language Resources and Evaluation},
  year      = {2008},
  volume    = {42},
  number    = {4},
  pages     = {335--359},
  doi       = {10.1007/s10579-008-9076-6}
}

@inproceedings{schmidt2018wesad,
  author    = {Schmidt, Philip and Reiss, Attila and Duerichen, Robert and
               Marberger, Claus and Van Laerhoven, Kristof},
  title     = {{WESAD}: A Multimodal Dataset for Wearable Stress and
               Affect Detection},
  booktitle = {Proc.\ 20th ACM Int.\ Conf.\ on Multimodal Interaction (ICMI)},
  year      = {2018},
  pages     = {400--408},
  doi       = {10.1145/3242969.3242985}
}

@article{cabrera2018matchnmingle,
  author    = {Cabrera-Quiros, Laura and Demetriou, Andrew and Balog, Egor
               and van der Meijden, Astrid and Gedik, Esma and
               Gebre, Binyam~G. and Hung, Hayley},
  title     = {The {MatchNMingle} Dataset: A Novel Multi-Sensor Resource for
               the Analysis of Social Interactions and Nonverbal Communication
               in Unstructured Mingle and Speed-Dating Scenarios},
  journal   = {IEEE Transactions on Affective Computing},
  year      = {2021},
  volume    = {12},
  number    = {1},
  pages     = {148--164},
  doi       = {10.1109/TAFFC.2018.2833645}
}

@article{sanchez2012emergent,
  author    = {Sanchez-Cortes, Dairazalia and Aran, Oya and
               Mast, Marianne~Schmid and Gatica-Perez, Daniel},
  title     = {A Nonverbal Behavior Approach to Identify Emergent Leaders
               in Small Groups},
  journal   = {IEEE Transactions on Multimedia},
  year      = {2012},
  volume    = {14},
  number    = {3},
  pages     = {816--832},
  doi       = {10.1109/TMM.2011.2181941}
}

@article{russell1980circumplex,
  author    = {Russell, James~A.},
  title     = {A Circumplex Model of Affect},
  journal   = {Journal of Personality and Social Psychology},
  year      = {1980},
  volume    = {39},
  number    = {6},
  pages     = {1161--1178},
  doi       = {10.1037/h0077714}
}

@article{mehrabian1996pleasure,
  author    = {Mehrabian, Albert},
  title     = {Pleasure-Arousal-Dominance: A General Framework for Describing
               and Measuring Individual Differences in Temperament},
  journal   = {Current Psychology},
  year      = {1996},
  volume    = {14},
  number    = {4},
  pages     = {261--292},
  doi       = {10.1007/BF02686918}
}

@article{bradley1994sam,
  author    = {Bradley, Margaret~M. and Lang, Peter~J.},
  title     = {Measuring Emotion: The Self-Assessment Manikin and the
               Semantic Differential},
  journal   = {Journal of Behavior Therapy and Experimental Psychiatry},
  year      = {1994},
  volume    = {25},
  number    = {1},
  pages     = {49--59},
  doi       = {10.1016/0005-7916(94)90063-9}
}

@incollection{john1999bigfive,
  author    = {John, Oliver~P. and Srivastava, Sanjay},
  title     = {The Big Five Trait Taxonomy: History, Measurement, and
               Theoretical Perspectives},
  booktitle = {Handbook of Personality: Theory and Research},
  edition   = {2nd},
  editor    = {Pervin, Lawrence~A. and John, Oliver~P.},
  publisher = {Guilford Press},
  year      = {1999},
  pages     = {102--138}
}

@article{watson1992traits,
  author    = {Watson, David and Clark, Lee~A.},
  title     = {On Traits and Temperament: General and Specific Factors of
               Emotional Experience and Their Relation to the Five-Factor Model},
  journal   = {Journal of Personality},
  year      = {1992},
  volume    = {60},
  number    = {2},
  pages     = {441--476},
  doi       = {10.1111/j.1467-6494.1992.tb00980.x}
}

@article{meng2026moderating,
  author    = {Meng, Xianxin and Li, Wei},
  title     = {Moderating Roles of the Big Five in Valence--Arousal Dynamics:
               A {TFace-Bi-GRU-SE} and {CTSEM} Study},
  journal   = {Information},
  year      = {2026},
  volume    = {17},
  number    = {5},
  pages     = {256},
  publisher = {MDPI},
  doi       = {10.3390/info17050256}
}

@book{rasmussen2006gaussian,
  author    = {Rasmussen, Carl~E. and Williams, Christopher~K.~I.},
  title     = {Gaussian Processes for Machine Learning},
  publisher = {MIT Press},
  year      = {2006},
  isbn      = {978-0-262-18253-9}
}

@inproceedings{um2017data,
  author    = {Um, Terry~T. and Pfister, Franz~M.~J. and Pichler, Daniel and
               Endo, Satoshi and Lang, Muriel and Bhatt, Didier and
               Fietzek, Urban and Kuli{\'c}, Dana},
  title     = {Data Augmentation of Wearable Sensor Data for {Parkinson's
               Disease} Monitoring Using Convolutional Neural Networks},
  booktitle = {Proc.\ 19th ACM Int.\ Conf.\ on Multimodal Interaction (ICMI)},
  year      = {2017},
  pages     = {216--220},
  doi       = {10.1145/3136755.3136817}
}

@inproceedings{eldele2021tscc,
  author    = {Eldele, Emadeldeen and Ragab, Mohamed and Chen, Zhenghua and
               Wu, Min and Kwoh, Chee~Keong and Li, Xiaoli and Guan, Cuntai},
  title     = {Time-Series Representation Learning via Temporal and
               Contextual Contrasting},
  booktitle = {Proc.\ 30th Int.\ Joint Conf.\ on Artificial Intelligence
               (IJCAI)},
  year      = {2021},
  pages     = {2352--2359},
  doi       = {10.24963/ijcai.2021/324}
}

@article{makowski2021neurokit2,
  author    = {Makowski, Dominique and Pham, Tam and Lau, Zen~Juen and
               Brammer, Jan~C. and Lespinasse, Fran{\c{c}}ois and
               Pham, Hung and Sch{\"o}lzel, Christopher and Chen, S.~H.~A.},
  title     = {{NeuroKit2}: A Python Toolbox for Neurophysiological Signal
               Processing},
  journal   = {Behavior Research Methods},
  year      = {2021},
  volume    = {53},
  number    = {4},
  pages     = {1689--1696},
  doi       = {10.3758/s13428-020-01516-y}
}

@article{greco2016cvxeda,
  author    = {Greco, Alberto and Valenza, Gaetano and Lanata, Antonio and
               Scilingo, Enzo~Pasquale and Citi, Luca},
  title     = {{cvxEDA}: A Convex Optimization Approach to Electrodermal
               Activity Processing},
  journal   = {IEEE Transactions on Biomedical Engineering},
  year      = {2016},
  volume    = {63},
  number    = {4},
  pages     = {797--804},
  doi       = {10.1109/TBME.2015.2474131}
}

@article{soleymani2012mahnob,
  author    = {Soleymani, Mohammad and Lichtenauer, Jeroen and Pun, Thierry
               and Pantic, Maja},
  title     = {A Multimodal Database for Affect Recognition and Implicit
               Tagging},
  journal   = {IEEE Transactions on Affective Computing},
  year      = {2012},
  volume    = {3},
  number    = {1},
  pages     = {42--55},
  doi       = {10.1109/T-AFFC.2011.25}
}

@book{hatfield1993emotional,
  author    = {Hatfield, Elaine and Cacioppo, John~T. and Rapson, Richard~L.},
  title     = {Emotional Contagion},
  publisher = {Cambridge University Press},
  year      = {1993},
  isbn      = {978-0-521-44948-3}
}

@inproceedings{lee2013pseudo,
  author    = {Lee, Dong-Hyun},
  title     = {Pseudo-Label: The Simple and Efficient Semi-Supervised
               Learning Method for Deep Neural Networks},
  booktitle = {Workshop on Challenges in Representation Learning,
               ICML},
  year      = {2013},
  volume    = {3},
  pages     = {896}
}

@article{subramanian2018ascertain,
  author    = {Subramanian, Ramanathan and Wache, Julia and Abadi, Mojtaba
               Khomami and Vieriu, Radu~L. and Winkler, Stefan and
               Sebe, Nicu},
  title     = {{ASCERTAIN}: Emotion and Personality Recognition Using
               Commercial Sensors},
  journal   = {IEEE Transactions on Affective Computing},
  year      = {2018},
  volume    = {9},
  number    = {2},
  pages     = {147--160},
  doi       = {10.1109/TAFFC.2016.2625250}
}

@article{mirandacorrea2021amigos,
  author    = {Miranda-Correa, Juan~Abdon and Abadi, Mojtaba Khomami and
               Sebe, Nicu and Patras, Ioannis},
  title     = {{AMIGOS}: A Dataset for Affect, Personality and Mood Research
               on Individuals and Groups},
  journal   = {IEEE Transactions on Affective Computing},
  year      = {2021},
  volume    = {12},
  number    = {2},
  pages     = {479--493},
  doi       = {10.1109/TAFFC.2018.2884461}
}

@article{park2020kemocon,
  author    = {Park, Cheul~Young and Cha, Narae and Kang, Soowon and
               Kim, Auk and Khandoker, Ahsan~Habib and
               Hadjileontiadis, Leontios and Oh, Alice and Jeong, Yong and
               Lee, Uichin},
  title     = {{K-EmoCon}: A Multimodal Sensor Dataset for Continuous Emotion
               Recognition in Naturalistic Conversations},
  journal   = {Scientific Data},
  year      = {2020},
  volume    = {7},
  number    = {1},
  pages     = {293},
  doi       = {10.1038/s41597-020-00630-y}
}

@article{prabhu2024collective,
  author    = {Prabhu, Navin Raj and Tsfasman, Maria and Oertel, Catharine
               and Gerkmann, Timo and Lehmann-Willenbrock, Nale},
  title     = {Dynamics of Collective Group Affect: Group-level Annotations
               and the Multimodal Modeling of Convergence and Divergence},
  journal   = {arXiv preprint arXiv:2409.08578},
  year      = {2024}
}

@inproceedings{mccowan2005ami,
  author    = {McCowan, Iain and Carletta, Jean and Kraaij, Wessel and
               Ashby, Simone and Bourban, S{\'e}bastien and Flynn, Mike and
               Guillemot, Ma{\"e}l and Hain, Thomas and Kadlec, Jaroslav and
               Karaiskos, Vasilis and Kronenthal, Melissa and
               Lathoud, Guillaume and Lincoln, Mike and Lisowska, Agnes and
               Post, Wilfried and Reidsma, Dennis and Wellner, Pierre},
  title     = {The {AMI} Meeting Corpus},
  booktitle = {Proc.\ 5th Int.\ Conf.\ on Methods and Techniques in
               Behavioral Research},
  year      = {2005}
}

@misc{alameda2015salsa,
  author        = {Alameda-Pineda, Xavier and Staiano, Jacopo and
                   Subramanian, Ramanathan and Batrinca, Ligia and
                   Ricci, Elisa and Lepri, Bruno and Lanz, Oswald and
                   Sebe, Nicu},
  title         = {{SALSA}: A Novel Dataset for Multimodal Group Behavior
                   Analysis},
  year          = {2015},
  eprint        = {1506.06882},
  archivePrefix = {arXiv},
  primaryClass  = {cs.CV}
}

@article{shu2018review,
  author    = {Shu, Lin and Xie, Jinyan and Yang, Mingyue and Li, Ziyi and
               Li, Zhenqi and Liao, Dan and Xu, Xiangmin and Yang, Xinyi},
  title     = {A Review of Emotion Recognition Using Physiological Signals},
  journal   = {Sensors},
  year      = {2018},
  volume    = {18},
  number    = {7},
  pages     = {2074},
  doi       = {10.3390/s18072074}
}

@article{ahmad2022survey,
  author    = {Ahmad, Zeeshan and Khan, Naimul},
  title     = {A Survey on Physiological Signal-Based Emotion Recognition},
  journal   = {Bioengineering},
  year      = {2022},
  volume    = {9},
  number    = {11},
  pages     = {688},
  doi       = {10.3390/bioengineering9110688}
}

@article{ventura2022validation,
  author    = {Bragan{\c{c}}a, Hendrio and Colonna, Juan~G. and
               Oliveira, Hor{\'a}cio~A.~B.~F. and Souto, Eduardo},
  title     = {How Validation Methodology Influences Human Activity
               Recognition Mobile Systems},
  journal   = {Sensors},
  year      = {2022},
  volume    = {22},
  number    = {6},
  pages     = {2360},
  doi       = {10.3390/s22062360}
}

@article{kreibig2010autonomic,
  author    = {Kreibig, Sylvia~D.},
  title     = {Autonomic Nervous System Activity in Emotion: A Review},
  journal   = {Biological Psychology},
  year      = {2010},
  volume    = {84},
  number    = {3},
  pages     = {394--421},
  doi       = {10.1016/j.biopsycho.2010.03.010}
}

@article{laeng2012pupillometry,
  author    = {Laeng, Bruno and Sirois, Sylvain and Gredeb{\"a}ck, Gustaf},
  title     = {Pupillometry: A Window to the Preconscious?},
  journal   = {Perspectives on Psychological Science},
  year      = {2012},
  volume    = {7},
  number    = {1},
  pages     = {18--27},
  doi       = {10.1177/1745691611427305}
}

@article{he2009imbalanced,
  author    = {He, Haibo and Garcia, Edwardo~A.},
  title     = {Learning from Imbalanced Data},
  journal   = {IEEE Transactions on Knowledge and Data Engineering},
  year      = {2009},
  volume    = {21},
  number    = {9},
  pages     = {1263--1284},
  doi       = {10.1109/TKDE.2008.239}
}

@article{johnson2019classimbalance,
  author    = {Johnson, Justin~M. and Khoshgoftaar, Taghi~M.},
  title     = {Survey on Deep Learning with Class Imbalance},
  journal   = {Journal of Big Data},
  year      = {2019},
  volume    = {6},
  number    = {1},
  pages     = {27},
  doi       = {10.1186/s40537-019-0192-5}
}

@article{geng2016label,
  author    = {Geng, Xin},
  title     = {Label Distribution Learning},
  journal   = {IEEE Transactions on Knowledge and Data Engineering},
  year      = {2016},
  volume    = {28},
  number    = {7},
  pages     = {1734--1748},
  doi       = {10.1109/TKDE.2016.2545658}
}

@inproceedings{wen2023ordinal,
  author    = {Wen, Changsong and Zhang, Xin and Yao, Xingxu and
               Yang, Jufeng},
  title     = {Ordinal Label Distribution Learning},
  booktitle = {Proc.\ IEEE/CVF Int.\ Conf.\ on Computer Vision (ICCV)},
  year      = {2023},
  pages     = {23481--23491}
}

@inproceedings{sohn2020fixmatch,
  author    = {Sohn, Kihyuk and Berthelot, David and Carlini, Nicholas and
               Zhang, Zizhao and Zhang, Han and Raffel, Colin~A. and
               Cubuk, Ekin~D. and Kurakin, Alexey and Li, Chun-Liang},
  title     = {{FixMatch}: Simplifying Semi-Supervised Learning with
               Consistency and Confidence},
  booktitle = {Advances in Neural Information Processing Systems},
  year      = {2020},
  volume    = {33},
  pages     = {596--608}
}

@article{iwana2021augmentation,
  author    = {Iwana, Brian~Kenji and Uchida, Seiichi},
  title     = {An Empirical Survey of Data Augmentation for Time Series
               Classification with Neural Networks},
  journal   = {PLOS ONE},
  year      = {2021},
  volume    = {16},
  number    = {7},
  pages     = {e0254841},
  doi       = {10.1371/journal.pone.0254841}
}

@inproceedings{tarvainen2017mean,
  author    = {Tarvainen, Antti and Valpola, Harri},
  title     = {Mean Teachers Are Better Role Models: Weight-Averaged
               Consistency Targets Improve Semi-Supervised Deep Learning
               Results},
  booktitle = {Advances in Neural Information Processing Systems},
  year      = {2017},
  volume    = {30}
}

@inproceedings{atcheson2017gp,
  author    = {Atcheson, Mia and Sethu, Vidhyasaharan and Epps, Julien},
  title     = {Gaussian Process Regression for Continuous Emotion Recognition
               with Global Temporal Invariance},
  booktitle = {Proc.\ IJCAI Workshop on Artificial Intelligence in Affective
               Computing},
  year      = {2017},
  series    = {Proceedings of Machine Learning Research},
  volume    = {66},
  pages     = {34--44},
  url       = {https://proceedings.mlr.press/v66/atcheson17a.html}
}

@inproceedings{seikavandi2025mumtaffect,
  author    = {Seikavandi, Meisam Jamshidi and Narcizo, Fabricio Batista and
               Vucurevich, Ted and Dittberner, Andrew Burke and Burelli, Paolo},
  title     = {{MuMTAffect}: A Multimodal Multitask Affective Framework for
               Personality and Emotion Recognition from Physiological Signals},
  booktitle = {Proceedings of the 3rd International Workshop on Multimodal and
               Responsible Affective Computing},
  pages     = {100--108},
  year      = {2025},
  publisher = {Association for Computing Machinery},
  doi       = {10.1145/3746270.3760232}
}

@inproceedings{seikavandi2025modelling,
  author    = {Seikavandi, Meisam J. and Fimland, Jostein and
               Narcizo, Fabricio Batista and Barrett, Maria and
               Vucurevich, Ted and Boldt, Jesper B{\"u}nsow and
               Dittberner, Andrew Burke and Burelli, Paolo},
  title     = {Modelling the Interplay of Eye-Tracking Temporal Dynamics and
               Personality for Emotion Detection in Face-to-Face Settings},
  booktitle = {BMVC 2025 MPI Workshop},
  publisher = {British Machine Vision Association},
  year      = {2025},
  url       = {https://bmva-archive.org.uk/bmvc/2025/assets/workshops/MPI/Paper_2/paper.pdf}
}

@article{seikavandi2026affectai,
  author  = {Seikavandi, Meisam Jamshidi and Modica, Alice and Obara, Anna and
             Narcizo, Fabricio Batista and Ignatenko, Tanya and
             Vucurevich, Ted and Boldt, Jesper B{\"u}nsow and Burelli, Paolo and
             Dittberner, Andrew Burke},
  title   = {{AffectAI-Capture}: A Reproducible Multimodal Protocol for
             Small-Group Meeting Research},
  journal = {arXiv preprint arXiv:2605.19794},
  year    = {2026},
  doi     = {10.48550/arXiv.2605.19794}
}

@article{seikavandi2026advancing,
  author    = {Seikavandi, Meisam Jamshidi and Dixen, Laurits and
               Fimland, Jostein and Desu, Sree Keerthi and
               Zserai, Antonia-Bianca and Lee, Ye Sul and Barrett, Maria and
               Burelli, Paolo},
  title     = {Advancing Face-to-Face Emotion Communication: A Multimodal
               Dataset ({AFFEC})},
  journal   = {IEEE Transactions on Affective Computing},
  year      = {2026},
  publisher = {IEEE},
  doi       = {10.1109/TAFFC.2026.3710632}
}

\end{document}